%% file: FRED_CameraReady.tex
\documentclass[letterpaper]{article} 
\usepackage{aaai24}  
\usepackage{times}  
\usepackage{helvet}  
\usepackage{courier}  
\usepackage[hyphens]{url}  
\usepackage{graphicx} 
\usepackage{natbib}  
\usepackage{caption} 
\usepackage{algorithm}
\usepackage{algorithmic}

\usepackage{newfloat}
\usepackage{listings}
\DeclareCaptionStyle{ruled}{labelfont=normalfont,labelsep=colon,strut=off} 
\floatstyle{ruled}
\newfloat{listing}{tb}{lst}{}
\floatname{listing}{Listing}
\usepackage{booktabs}
\usepackage{tabularx}
\usepackage{graphicx}
\usepackage{multirow}
\usepackage{subcaption}
\usepackage{nicefrac}       
\usepackage{amsmath}
\usepackage{amssymb}
\usepackage{colortbl}
\usepackage[noabbrev, capitalize]{cleveref}

\title{FRED: Towards a Full Rotation-Equivariance in Aerial Image Object Detection}
\author{
    Chanho Lee\textsuperscript{\rm 1},
    Jinsu Son\textsuperscript{\rm 1},
    Hyounguk Shon\textsuperscript{\rm 1},
    Yunho Jeon\textsuperscript{\rm 2},
    Junmo Kim\textsuperscript{\rm 1}
}
\affiliations{
    \textsuperscript{\rm 1}Korea Advanced Institute of Science and Technology, South Korea\\
    \textsuperscript{\rm 2}Hanbat National University, South Korea\\
    \{yiwan99, sonjs, hyounguk.shon\}@kaist.ac.kr, yhjeon@hanbat.ac.kr, junmo.kim@kaist.ac.kr

}

\usepackage{bibentry}

\begin{document}

\maketitle

\begin{abstract}
Rotation-equivariance is an essential yet challenging property in oriented object detection. While general object detectors naturally leverage robustness to spatial shifts due to the translation-equivariance of the conventional CNNs, achieving rotation-equivariance remains an elusive goal. Current detectors deploy various alignment techniques to derive rotation-invariant features, but still rely on high capacity models and heavy data augmentation with all possible rotations. In this paper, we introduce a Fully Rotation-Equivariant Oriented Object Detector (FRED), whose entire process from the image to the bounding box prediction is strictly equivariant. Specifically, we decouple the invariant task (object classification) and the equivariant task (object localization) to achieve end-to-end equivariance. We represent the bounding box as a set of rotation-equivariant vectors to implement rotation-equivariant localization. Moreover, we utilized these rotation-equivariant vectors as offsets in the deformable convolution, thereby enhancing the existing advantages of spatial adaptation. Leveraging full rotation-equivariance, our FRED demonstrates higher robustness to image-level rotation compared to existing methods. Furthermore, we show that FRED is one step closer to non-axis aligned learning through our experiments. Compared to state-of-the-art methods, our proposed method delivers comparable performance on DOTA-v1.0 and outperforms by 1.5 mAP on DOTA-v1.5, all while significantly reducing the model parameters to 16\%.
\end{abstract}

\section{Introduction}
Aerial object detection is an emerging field in the domain of computer vision. Since aerial images capture objects with arbitrary orientations and are often densely packed, oriented bounding box (OBB) can provide a tighter representation in such cases. One distinguishing feature of aerial images is the non-axis aligned nature, absence of any top-bottom or left-right bias. The ideal aerial object detector should consistently deliver predictions irrespective of object orientation. If the image undergoes rotation, the predicted OBB should also rotate concurrently. Hence, for an oriented object detector to be reliable, it must exhibit rotation-equivariance.
\begin{figure}[t]
\centering
\includegraphics[width=\linewidth]{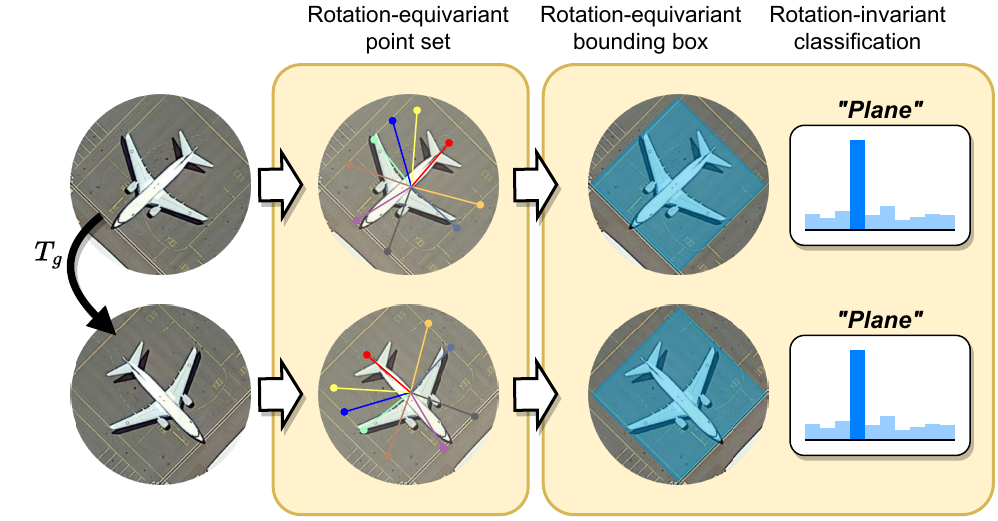}
\caption{\textbf{Overview of the fully rotation-equivariant object detector (FRED).} FRED consists of a rotation-equivariant backbone which predicts a point set followed by two prediction branches -- (1) a rotation-equivariant box regression head and (2) a rotation-invariant classification head. We achieve end-to-end equivariance for object detection.} 
\label{fig:figure1}
\end{figure}

\begin{figure*}[t]
\centering
\includegraphics[width=0.8\textwidth]{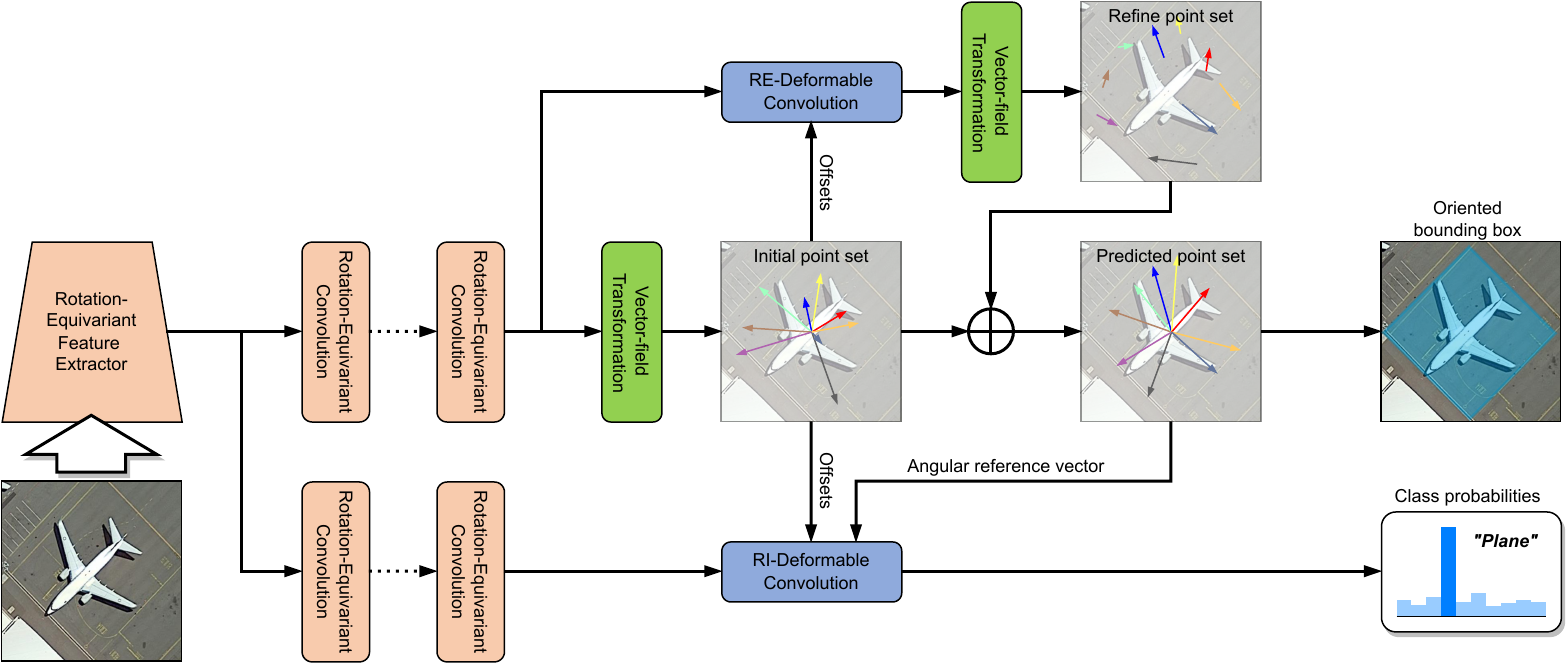} 
\caption{\textbf{Overall model architecture of the proposed Fully Rotation-Equivariant Detector (FRED).} \( C_N \)-equivariant features are fed into the rotation-equivariant head up to two deformable convolution blocks. The Rotation-Equivariant Deformable Convolution (RE-DCN) tilizes an initial point set as an offset and refines it through spatial adaptation without breaking rotation-equivariance. The Rotation-Invariant Deformable Convolution (RI-DCN) performs an orientation alignment to produce rotation-invariant features using an align reference vector sourced from the localization branch. As both the deformable offsets and the reference vector maintain rotation-equivariance, the classification branch achieves instance-level rotation-invariance.}
\label{fig:figure2}
\end{figure*}
However, achieving rotation-equivariance on oriented object detection is challenging, since most researches are extended from horizontal object detection models. Most methods rely on the assumption of accurate orientation estimation and focus on making features invariant to rotation. A representative approach to achieve this is the RoI Transformer \citep{roitransformer}, which leverages rotation-sensitive region-of-interest (RoI) pooling on a rotated RoI (RRoI) to acquire instance-level rotation-invariance. Such orientation-specific feature refinement has demonstrated its efficiency across one-stage object detectors \citep{han2021s2anet} and anchor-free detectors \citep{pan2020drn}. Another strategy is using point set representation which implicitly represents the oriented bounding box as a set of adaptively learned points \citep{guo2021beyond, li2022oriented}. These approaches have managed to separate out the prediction of orientation itself, naturally leads to non-axis aligned feature learning. Yet, these aforementioned methods rely heavily on data augmentation using random rotations and remain distant from achieving true rotation-equivariance.

Recently, \citet{han2021ReDet} proposed ReDet, a rotation-equivariant detector firstly employing rotation-equivariant CNNs \citep{weiler2019general}. Their Rotation-invariant ROI Align (RiRoI Align) leverages the characteristics of rotation-equivariant features, enabling the extraction of rotation-invariant features dependent on the rotated RoI. However, it is worth noting that even though their RiRoI Align operates based on a rotation-equivariant theory, the orientation of the predicted RRoI itself is not equivariant. Due to the ambiguity and angular discontinuity of OBB representations, rotation-invariance of RiRoI Align is vulnerable to significant degrees of rotations. To design a model that remains consistent across any rotation, we utilize a point set representation in place of the bounding box, and endow the entire localization process with strict rotation-equivariance.

In this work, we propose a fully rotation-equivariant oriented object detector named \emph{FRED} which leverages point set representation to achieve full rotation-equivariance on both classification and localization. We conceptualize the bounding box as a set of rotation-equivariant vectors. By employing this idea, we ensure that with any rotation, the vectors not only shift in accordance with the image-level rotation but also change their orientation simultaneously. This trait perfectly satisfies the attributes needed for oriented bounding box prediction as depicted in \cref{fig:figure1}. Furthermore, we apply these rotation-equivariant vectors as offsets of deformable convolution. This allows us to propose Rotation-Equivariant Deformable Convolution (RE-DCN) and Rotation-Invariant DCN (RI-DCN), which can simultaneously achieve spatial and orientation alignment through an rotation-equivariant receptive field. Compared to previous methods, our FRED is highly robust to image rotations powered by end-to-end rotation-equivariance. Moreover, FRED maximizes the benefits of the high-level weight sharing of rotation-equivariant CNNs, showcasing superior performance with fewer learnable parameters than any other detectors.

Moreover, we discovered a promising phenomenon while training our rotation-equivariant model. Typically, rotation-equivariance is associated with robustness to different rotations for a single instance. If there exists an instance group with similar context and scale, we can anticipate rotation-equivariance among them. We observed that FRED, just before full convergence, learned the relative pose of objects without any direction-specific supervision. While this tendency diminishes during the convergence process, it can be seen as a reflection of FRED being trained in a genuinely non-axis aligned manner. Through our experiments, we demonstrate that previous non-axis aligned methods are still being trained in an axis-overfitted manner, while our FRED showcases a more genuine non-axis aligned learning.

In summary, our main contributions in this paper are as follows:
\begin{itemize}
    \item To the best of our knowledge, we are the first to propose a fully rotation-equivariant oriented object detector. Compared to previous state-of-the-art methods, FRED guarantees more robust predictions against image rotations.
    \item We propose novel methods that combine deformable convolution and rotation-equivariant vectors to simultaneously perform spatial and orientation alignment without disrupting equivariance.
    \item Our experiments demonstrate that FRED achieves promising results with significantly fewer parameters, and offers a new insight into axis-free learning.
\end{itemize}

\section{Related Works}

\subsection{Oriented Object Detection}
The main approach for oriented object detection extends from horizontal object detection with additional regression for orientation. Challenges in oriented object detection in aerial images arise from arbitrary oriented and densely packed objects. RoI Transformer \citep{roitransformer} proposed a rotation-sensitive RoI pooling for obtaining rotation-invariance. SCRDet \citep{scrdet}, DRN \citep{pan2020drn}, S2A-Net \citep{han2021s2anet}, and R$^3$Det \citep{yang2021r3det} adressessed the challenges through methods that refine features. Most research focuses on developing methods to apply the axis-aligned property to non-axis aligned oriented object detection, yet rely on well predefined anchors and angular discontinuity of OBB.\citep{qian2021rsdet, yang2022kfiou}

\subsection{Alternative Bounding Box Representations}
The introduction of point set representation offers a promising alternative to these issues, escaping from traditional anchors and bounding boxes. By adaptively capturing object context, point set has demonstrated their capability to learn richer representations as shown in RepPoints \citep{yang2019reppoints}. Through strategies like convex-hull feature adaptation \citep{guo2021beyond} and orientation-sensitive sampling \citep{li2022oriented}, point set based methods outperform the previous anchor-based detectors. Our FRED introduces rotation-equivariant point set prediction and its benefit, showcasing its closer alignment with true non-axis aligned learning.

\subsection{Rotation-Equivariant Neural Networks}
Beginning with the Group-Equivariant CNN proposed by \citet{cohen2016group}, several methods have introduced rotation-equivariant CNNs using steerable filters \citep{weiler2019general, cesa2021program} and they have been proven effective in various imagery fields \citep{veeling2018rotation, gupta2021rotation, lee2023learning}. Recently, \citet{han2021ReDet} introduced ReDet, a rotation-equivariant detector for oriented object detection that firstly utilizes the rotation-equivariant CNNs. While RiRoI Align offers a rotation-invariant transform, its dependency on the non-equivariantly predicted RRoI still introduces a residual challenge to achieving full rotation-equivariance.

\section{Preliminaries}
This section offers a brief overview on the concept of rotation-equivariance. Given the prevalent use of steerable filters to yield rotation-equivariant features, the finite rotation group and its representation through group-wise permutation are introduced.

Let \( G \) be a group which can be any transformations on image space \( X \). Then a function \( \Phi: X \to Y \) is said to be \textit{equivariant} if
\begin{equation}
     \Phi\left(T^X_g(x)\right) = T^Y_g\left(\Phi(x)\right)  \quad \forall g \in G ,\, \forall x \in X
\end{equation}
where \(T^X_g\) and \(T^Y_g\) is a group action defined on each space. If  \(T^Y_g\) is identity mapping, then invariance holds. For example, the conventional CNN \(\Phi\) shares convolution weight at every location, so satisfies the equation above on translation action.

\subsection{Rotation-Equivariance} 
In this paper, we are addressing rotation-equivariance, so group can be formulated as the semi-direct product of the translation group \( (R^2, +) \) and rotation group \( H \). For a group \(G\), the rotation-equivariance of function \(f: X \to Y\) can be expressed as
\begin{equation}
     f\left(T^X_g(x)\right) = T^Y_g\left(f(x)\right)  \quad \forall g \in G ,\, \forall x \in X
\end{equation}
 given \(T^X_g\) and \(T^Y_g\) as rotation action on \(X\) and \(Y\) respectively. If \(T^X_g\) and \(T^Y_g\) are isomorphic image-level rotations in each space, then function \(f\) can be viewed as transforming the image space into a rotation-equivariant scalar field.
 
On the other hand, let a function \(v: X \to V\) be a mapping from the image space \(X\) to 2-dimensional vector field space \(V\). To provide clearness in notation,  let us denote \(T_g\)  as the group action of \(G\) that operates on both \(X\) and \(V\). Then rotation-equivariance of vector field can be expressed as:
\begin{equation}
     v\left(T_g(x)\right) = R_g T_g\left(v(x)\right)  \quad \forall g \in G ,\, \forall x \in X
\end{equation}
In this context, \(R_g\) is a group action on \(G\) that rotates every vector in parallel with \(T_g\). For clarity, a rotation-equivariant vector field necessitates not just image-level rotations \(T_g\) but also ensures that the vector predicted at each image pixel rotates by \(R_g\).

\begin{figure*}[t]
    \begin{subfigure}{.5\textwidth}
    \centering
    \includegraphics[width=0.9\linewidth]{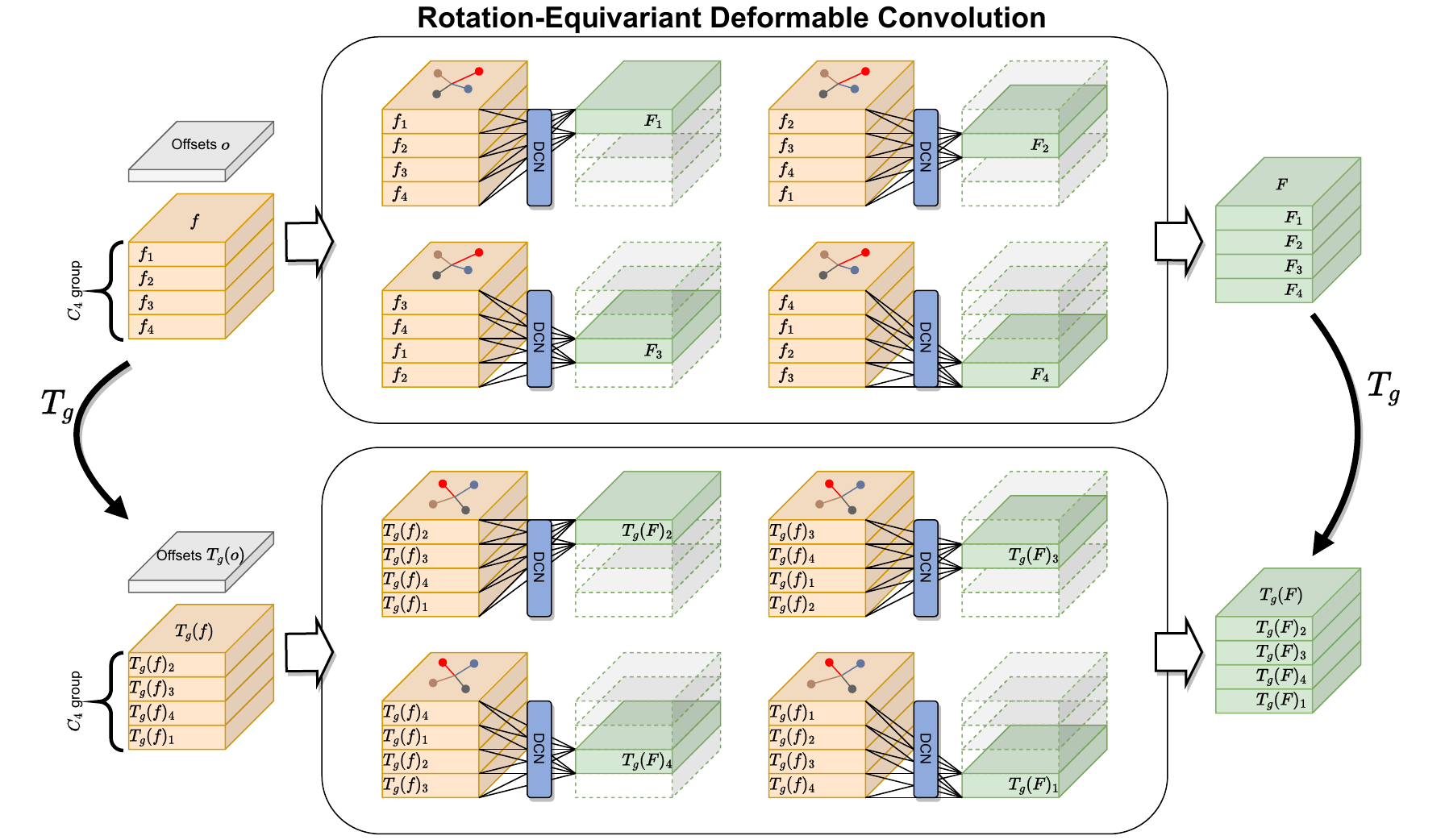}
    \caption{\textbf{Rotation-equivariant Deformable Convolution}}
    \label{fig:figure3a}
    \end{subfigure}
    \begin{subfigure}{.5\textwidth}
    \centering
    \includegraphics[width=0.9\linewidth]{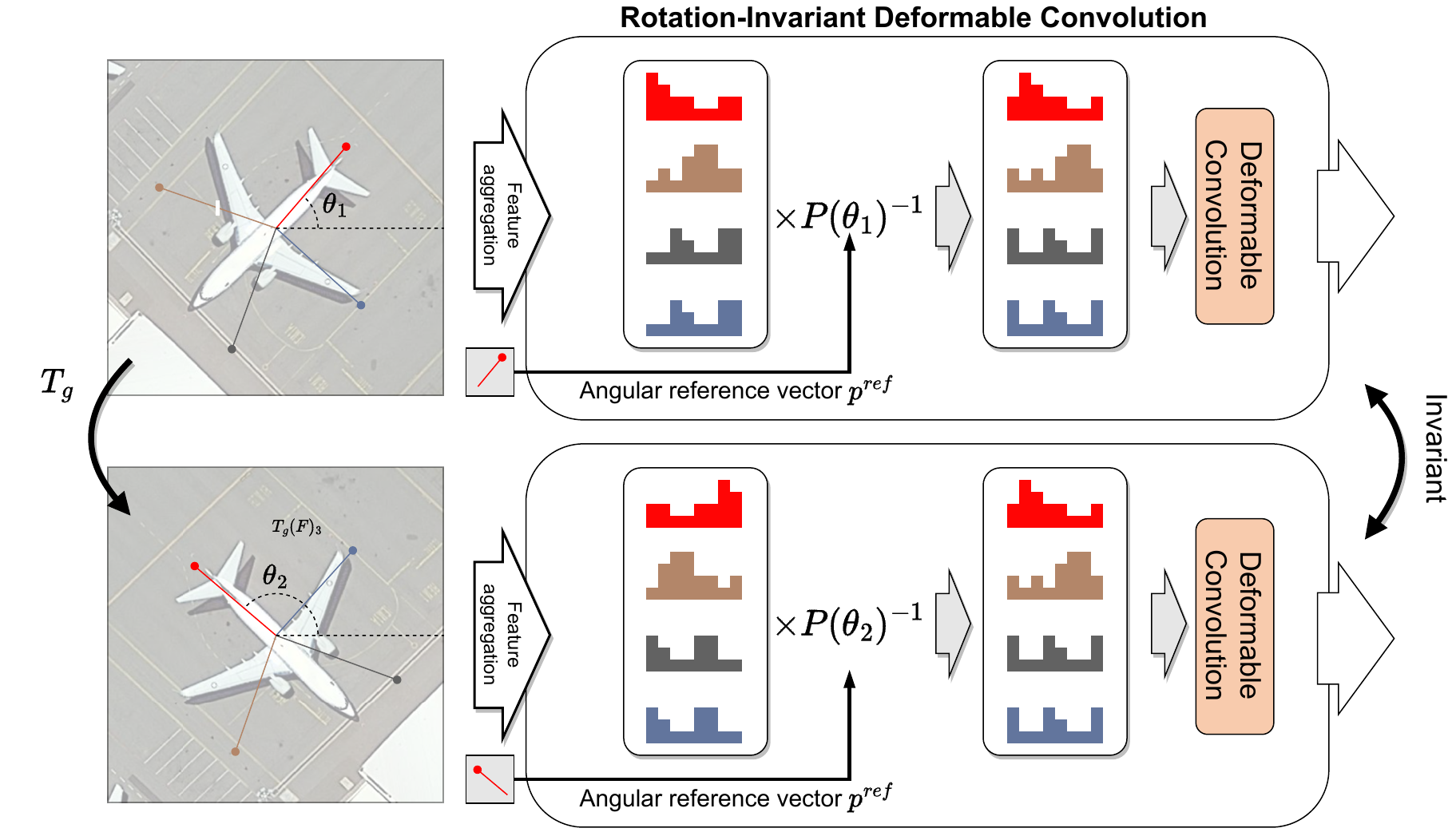}
    \caption{\textbf{Rotation-invariant Deformable Convolution}}
    \label{fig:figure3b}
    \end{subfigure}
\caption{This example illustrates a 4-equivariant rotation group ($C_4$) and 2x2 deformable kernel for simplicity. The deformable convolution (DCN) layer parameters are shared between rotation groups.}
\label{fig4}
\end{figure*}

\subsection{Steerable Filters and Cyclic-Equivariance} 
A commonly used method to implement a practical rotation-equivariant CNN is to utilize steerable filters. If a steerable filter rotates at intervals of \(2\pi/N\) degree and forms \(N\) weight-shared filters, we can create a convolution layer that is discretely rotation-equivariant to the cyclic group \(C_N\). For a group action \(T_g\) on the cyclic group \(C_N\), cyclic-equivariance can be expressed as:
\begin{equation}
     f\left(T_g(x)\right) = P_g T_g\left(f(x)\right)  \quad \forall g \in C_N ,\, \forall x \in X
\end{equation}
where \(P_g\) is rotation group-wise permutation operator. For example, when the image is rotated by \(2\pi n/N\), the \(C_N\)-equivariant feature undergoes both an image-level rotation and a group-wise permutation of degree \(n\). It's worth noting that every intermediate feature of a rotation-equivariant CNN is cyclic-equivariant. One major property of cyclic-equivariant feature is that it can be transformed into either rotation-invariant or rotation-sensitive features through a pooling operation. As in \citet{weiler2019general}, rotation-invariant features are obtained from the maximum response across all rotations, achieved through rotation group-wise max pooling such as \(\max_g f(x)\).

On the other hand, rotation-equivariant vectors can be obtained not just from the max response, but also using the argmax operator. By combining max response \(\max_g f(x)\) and its orientation \(\theta=\frac{2\pi}{N} \arg\max_g f(x)\), we can transform a cyclic-equivariant feature to a vector field \(v(x)\) as
\begin{align}
     v(x) = \left(\max_g f(x)\right) \cdot \left[\cos(\theta), \sin(\theta)\right]^T
     \label{eq:vec}
\end{align}

For distinguishing the rotation-equivariant vector field, we will call the rotation-equivariant scalar field as instance-level rotation-invariance.

\section{Methodology}
Initially, rotation-equivariant features are extracted from the rotation-equivariant backbone and neck. To ensure rotation-equivariant localization with the point set \( S = \{[x^k, y^k]^\top\}_{k=1}^K \), we transform cyclic-equivariant features into \(K\) vector fields as \cref{eq:vec}, through a vector-field transformation layer. Our core idea is to utilize this point set as an offset for deformable convolution, enabling a rotation-equivariant adaptive receptive field. In the localization branch, Rotation-Equivariant Deformable Convolution (RE-DCN) maintains cyclic-equivariance to perform localization refinement through spatial adaptation. On the other hand, the Rotation-Invariant Deformable Convolution (RI-DCN) transforms features into invariant ones to ensure robust classification. We have advanced the orientation alignment proposed in ReDet \citep{han2021ReDet}, obtaining better rotation-equivariance and strengthening the connection between localization and classification through our alignment reference vector driven by point set prediction. Finally, we introduce the edge constraint loss to assure orientation sensitivity and stable training of the alignment reference vector. The overall architecture of our proposed method is depicted in \cref{fig:figure2}. 

\subsection{Rotation-Equivariant Deformable Convolution} 
The key role of rotation-equivariant offset is to guarantee that each kernel of the deformable convolution consistently focuses on a semantically identical area. However, directly applying regular deformable convolution breaks rotation-equivariance since it blends the features of all rotation groups, being agnostic to their distinctions. To solve this, we introduce Rotation-Equivariant Deformable Convolution (RE-DCN) which allows all \(N\) rotation groups to perform independently. As shown in \cref{fig:figure3a}, each rotation group independently executes deformable convolution using shared offsets and weights. Through the use of RE-DCN, the output feature retains cyclic-equivariance, allowing it to be transformed into a rotation-equivariant vector field. This approach not only ensures spatial feature refinement without compromising rotation-sensitivity but also achieves an N-factor parameter efficiency through weight sharing. A detailed structure and the corresponding pseudo code are provided in the appendix.

\subsection{Rotation-Invariant Deformable Convolution}
The classification branch needs consistent prediction that comes from rotation-invariance, rather than retaining the feature's orientation. At a basic level, if we make the feature that is input to the deformable convolution rotation-invariant, the output can also achieve invariance. A potential approach to this is the rotation group-wise max pooling mentioned earlier \citep{cohen2016group}. However, this strategy leads to a significant reduction in the model's capacity due to a dramatic channel reduction, and it also sweeps away all sensitive orientation information.

To tackle this issue, ReDet \citep{han2021ReDet} introduced the orientation alignment, which can be formulated as:
\begin{equation}
    OA(f, \theta) = \textit{Int}\left(SC(f, r), \theta\right), \quad r=\lfloor \theta N/2\pi \rfloor
\end{equation}

where $\textit{Int}(\cdot)$ denotes bilinear interpolation between adjacent rotation groups and \(SC(\cdot)\) is rotation group-wise permutation. In simpler terms, orientation alignment can be seen as an inverse direction of cyclic permutation \(P_g\), aiming to achieve rotation-invariance. While the orientation alignment is itself a rotation-equivariant operator, it is the \(\theta\) that truly determines rotation-invariance.  It is important to highlight that the orientation alignment introduced in ReDet has certain shortcomings in ensuring rotation-invariance. Specifically, (a) the \(\theta\) is derived from the RoI head, which comprises regular CNNs, so eventually loses its rotation-equivariance, and (b) even if 
\(\theta\) is predicted with utmost accuracy, its efficacy still hinges on the specific representation style of the oriented bounding box and predefined anchors.

In contrast, each individual point of our predicted localization is rotation-equivariant, making it suitable for orientation alignment. Our approach ensures that (a) through our proposed localization branch, \(\theta\) naturally attains rotation-equivariance, (b) the network can adaptively learn the appropriate direction without constraints imposed by the OBB representation style, and (c) it strengthens the correlation between localization and classification. To ensure robustness to the feature grid location, we define the angular reference vector \(\vec{p}^{ref}\) for orientation alignment as \(\vec{p}^{ref}=[x^{ref}-x^c,y^{ref}-y^c]^\top\), where  \([x^{ref}, y^{ref}]^\top\) is the alignment reference point and \([x^c, y^c]^\top\) is the center point of the convex hull formed through the point set \( S = \{[x^k, y^k]^\top\}_{k=1}^K \), respectively. Then, our oriented alignment is defined as
$OA(f, \measuredangle \vec{p}^{ref} )$ where $\measuredangle(\cdot)$ is the angle from the x-axis. 

We emphasize the fact that our point set prediction is made up of rotation-equivariant vectors. This means any point within the set can serve as an alignment reference, ensuring rotation-invariance of alignment.
Through our proposed orientation alignment, the transformed rotation-invariant feature is fed into the deformable convolution and utilized for classification. We refer to this entire process as Rotation-Invariant Deformable Convolution (RI-DCN), which is depicted in \cref{fig:figure3a}. 

\subsection{Orientation Alignment without Degeneration}
Although every rotation-equivariant point can be utilized as an alignment reference, we found that randomly selecting a point can sometimes lead to significant training instability. In point set based methods \citep{yang2019reppoints, guo2021beyond, li2022oriented}, we observed that a few points might be always excluded from the convex-hull, and wander around the center of the object. Although such points still remain rotation-equivariant, their direction can change dramatically during training, potentially leading to noisy alignment and unstable learning. Therefore, without majorly impacting the existing localization performance, there is a need to ensure that the reference point is influential in both localization and the formation of the convex hull. We introduce the edge constraint loss, pushing the reference point towards the nearest center points of the four edges of the ground truth bounding box. Given the four center points from the ground truth box edges, denoted as \( \{[x^g_i, y^g_i]^\top\}_{i=1}^4 \), the edge constraint loss \( \mathcal{L}_{ec}\) can be formulated as
\begin{equation}
    \mathcal{L}_{ec} = \min_i \left\lVert [x^{ref}, y^{ref}]^\top - [x^g_i, y^g_i]^\top \right\rVert_2
\end{equation}
We set a default weight of edge constraint loss as 0.025 to minimize its influence on localization loss. Without losing generality, we can set the first point of the predicted point set as the alignment reference.

\begin{table*}[t]
   \input{tables/dota_sota.tex}
    \caption{\textbf{Comparisons with state-of-the-art methods on DOTA-v1.0 OBB Task}. Res50, Res101, H104, and ReRes50 mean ResNet50, ResNet101, Hourglass104, Rotation-equivariant ResNet50, respectively. The best and second-best results are boldfaced and underlined, respectively.}
   \label{tab:dota_sota}
\end{table*}

\begin{table*}[t]
   \input{tables/dota_v15_sota.tex}
    \caption{\textbf{Comparisons with state-of-the-art methods on DOTA-v1.5 OBB Task.} RetinaNet-O, Mask R-CNN, HTC, and ReDet reported from \citep{han2021ReDet}.
    }
   \label{tab:dotav15_sota}
\end{table*}

\section{Experiments}

\subsection{Benchmark and Implementation Details}
\textbf{DOTA dataset} \citep{DOTA,9560031} is a large-scale benchmark designed for assessing oriented object detection in aerial images. It includes 2,806 aerial images with sizes varying from $800\times800$ to $4000\times4000$. In \textbf{DOTA-v1.0}, there are 188,282 instances distributed among 15 specific categories: Plane (PL), Baseball diamond (BD), Bridge (BR), Ground track field (GTF), Small vehicle (SV), Large vehicle (LV), Ship (SH), Tennis court (TC), Basketball court (BC), Storage tank (ST), Soccer-ball field (SBF), Roundabout (RA), Harbor (HA), Swimming pool (SP), and Helicopter (HC). \textbf{DOTA-v1.5} shares the image set with DOTA-v1.0 but introduces an additional class: Container Crane (CC), bringing the total to 402,089 instances. With the inclusion of extremely small objects, DOTA-v1.5 is more challenging, yet it seems to provide more reliable labels. For experimental settings, both the training and validation sets from DOTA are combined for training, with the test set reserved for evaluations. Images are typically split into 1024×1024 patches using an 824 stride. For evaluation metric, the mean average precision (mAP) metric \citep{everingham2010pascal} is used.

\textbf{Model Architecture.} Our implementation is based on the MMRotate \citep{zhou2022mmrotate} and $E(2)$-CNN \citep{weiler2019general} framework. Our proposed model is based on \(C_8\) ReResNet-50 backbone pretrained on ImageNet, with ReFPN neck proposed in \citet{han2021ReDet}. We stacked three C8 rotation-equivariant convolution layers for each branch, and used modulated deformable convolution \citep{zhu2019deformable} for the classification branch. Focal loss \citep{lin2017focal}, convex IOU loss \citep{rezatofighi2019convexiou} and spatial constraint loss with APAA strategy as described in \citet{li2022oriented} is employed for training. We set the weight of our edge constraint loss as 0.0025 as default.

\textbf{Training Scheme.} The training was conducted with the stochastic gradient descent optimizer with the momentum and the weight decay set to $0.9$ and $0.0001$, respectively. The initial learning rate is 0.008, and the model is trained for 40 epochs with batch size 8, using a step decay schedule.To ensure a fair comparison, we refrained from fine-tuning any hyper-parameters related to losses, sampling strategies, and training schedules, in line with the settings from  \citet{guo2021beyond} and \citet{li2022oriented}.

\textbf{Strided Convolution and Equivariance.} Due to the discreteness of images, rotation-equivariant CNNs are strictly equivariant only for rotations that are multiples of 90 degree.  However, employing even-sized images and strided convolution layers can disrupt this strict equivariance, even for 90-degree rotations. This issue arises because strided convolution always samples the top-left corner, as reported by \citet{romero2020attentive}. While the impact on performance might be minimal, given our goal to achieve rotation robustness through not approximate but perfect rotation-equivariance, we decided to pursue a more stringent rotation-equivariance. To address this issue, by simply adding zero-padding to ensure strided convolution always experience odd-numbered images, strict equivariance can be achieved without any significant modifications to the network structure. A more detailed analysis of how strided convolution can disrupt rotation-equivariance is covered in the appendix.
\begin{figure}[t]
    \centering
\includegraphics[width=\linewidth]{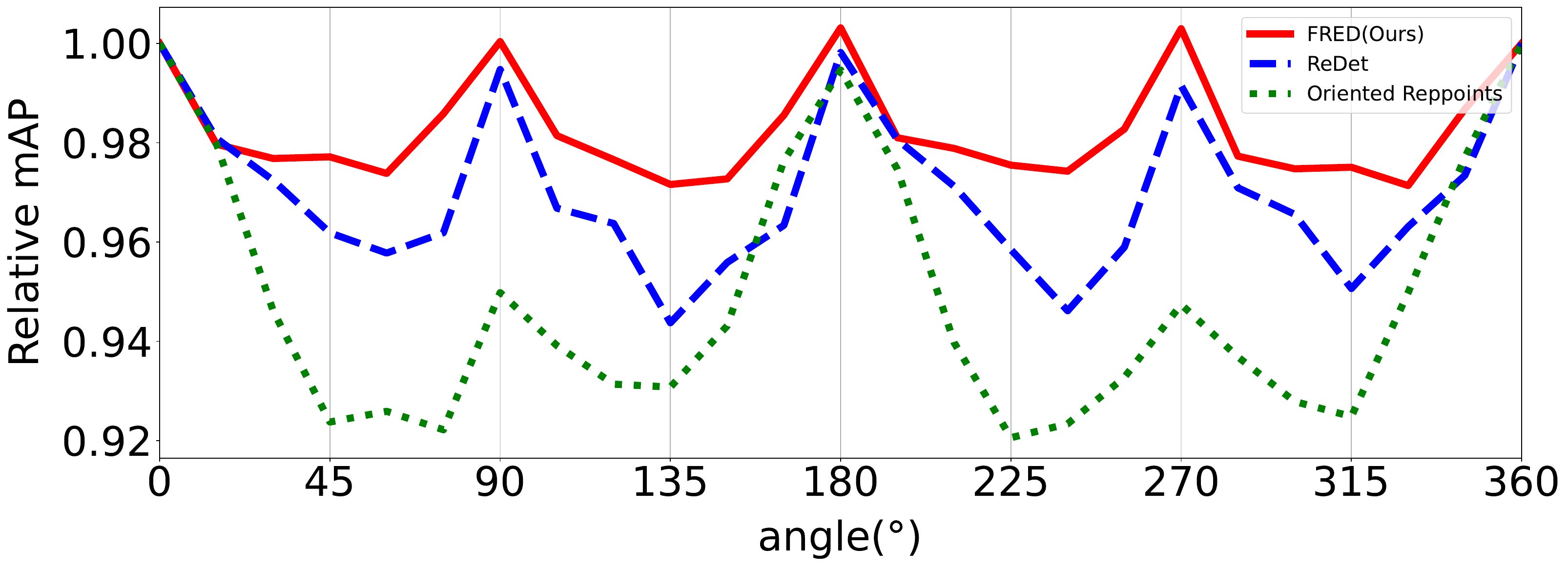}
\caption{\textbf{Robustness against rotation} estimated with DOTA-v1.0. We compared the performance degradation of various models as the image rotates. Excluding discrete rotations at 90-degree intervals, the loss of rotated image information always results a decreased mAP.}
\label{fig5}
\end{figure}

\begin{figure*}[t]
\centering
\includegraphics[width=0.95\textwidth]{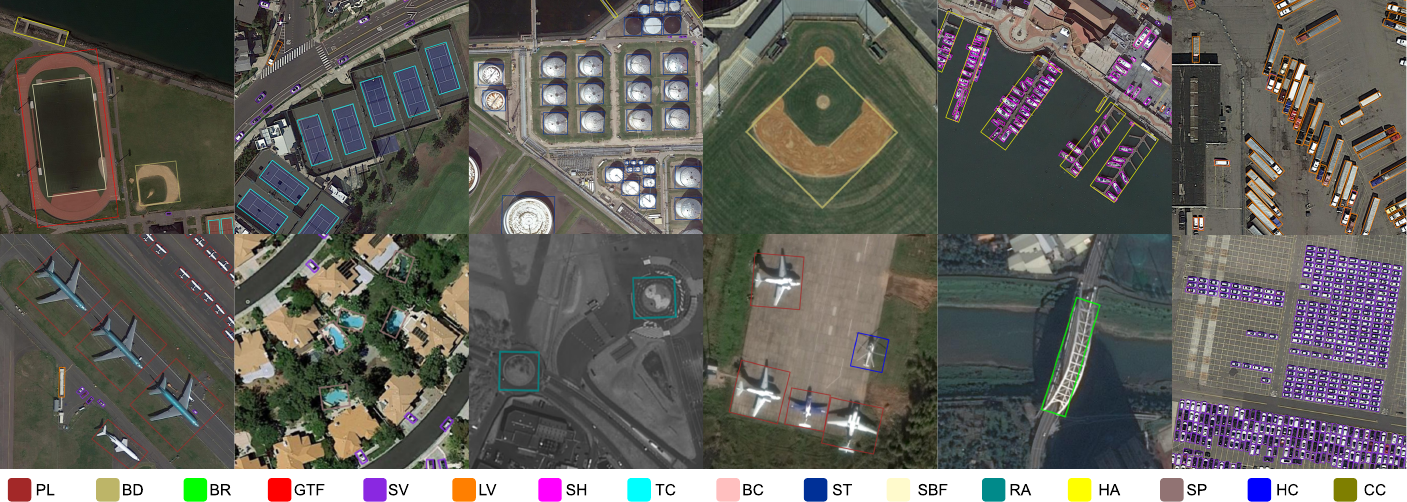} 
\caption{\textbf{Examples of detection results using FRED on DOTA-v1.5}}
\label{fig:figure6}
\end{figure*}

\begin{table}[t]
    \input{tables/ablation.tex}
\caption{\textbf{Ablations incrementally added equivariances.} ``Vector-field'' refers to vector-field transformation layer. }
\label{table:equivariance-ablation}
\end{table}

\begin{figure}[t]
    \centering
    \begin{subfigure}{.48\linewidth}
        \centering
        \includegraphics[width=\linewidth]{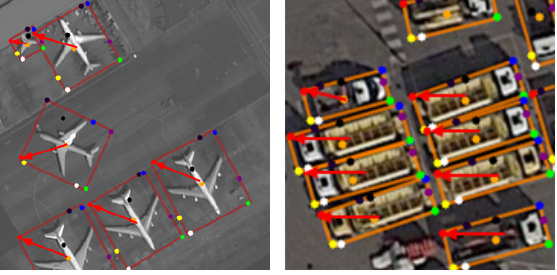}
        \caption{Oriented RepPoints}
        \label{fig:previous}
    \end{subfigure}
    \begin{subfigure}{.48\linewidth}
        \centering
        \includegraphics[width=\linewidth]{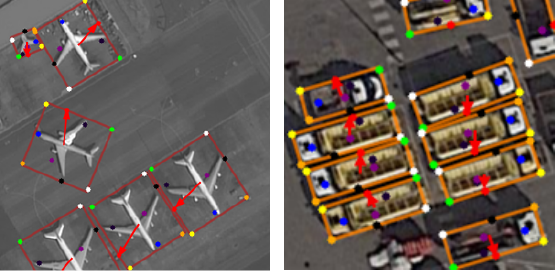}
        \caption{FRED (Ours)}
        \label{fig:proposed}
    \end{subfigure}
    \caption{\textbf{Directionality emerges with the equivariant point set representation.} Compared to Oriented RepPoints \citep{li2022oriented}, FRED maintains object orientation without explicit supervision. For visualization purposes, we color-coded each point in the predicted point set with respect to the ordering. Arrow indicates the first point in the set. Both models are trained for 30 epochs and early-stopped before convergence.}
    \label{fig:figure5}
\end{figure}

\subsection{Comparison with the State-of-the-art Methods}
\textbf{DOTA-v1.0.} Based on the experimental results on the single-scale DOTA-v1.0 dataset, we aim for a fair comparison with previous methods. Our model, using the ReResNet50 and ReFPN backbone, shows comparable result ranked second among anchor-free methods. we note that our model consists of only 16\% of model size.

\noindent\textbf{DOTA-v1.5.} Similar to DOTA-v1.0, we also report single-scale experiments for DOTA-v1.5. For a fair comparison, we reimplemented OrientedRepPoints using the official code. Our model achieved 78.3 mAP and surpasses the state-of-the-art anchor-free method by 1.4 mAP. Given that our model demonstrates superior results in more difficult and reliable dataset settings, we assert that it possesses a greater generalization capability.

\noindent\textbf{Robustness against rotations.} The goal of our fully rotation-equivariant model is to provide consistently reliable predictions during the inference phase regardless of rotations. We observed how the performance changes by varying the rotation of the test images. As seen in \cref{fig5}, while our method consistently performs regardless of image rotation, non-equivariant models exhibit significant variations in performance. Specifically, while the previous methods show performance degradation for large rotations, our model comparably maintains its performance. Even though our model is designed to be discretely rotation-equivariant at 45-degree intervals, it exhibits approximate rotation-equivariance across every continuous rotations. 

\section{Discussion}
\noindent\textbf{Ablation on rotation-equivariances.}
 To understand the effectiveness of rotation-equivariance for both object classification and localization, we conducted experiments on the DOTA-v1.0 dataset. To gauge the efficacy of rotation-equivariance on each branch, we measured performance by incrementally adding various rotation-equivariances based on the ReResNet50 and ReFPN feature extractor.
 
In \cref{table:equivariance-ablation}, the term 'Vector-field' refers to the implementation of a rotation-equivariant vector-field in the localization branch. GroupPool and Orientation Alignment (OA) are both rotation-invariance for classification branch, but with different way. The efficiency of rotation-equivariance for both classification and localization is evident. Without any equivariance transformation, a model simply becomes in a capacity reduced model with rotation-agnostic behavior. Since group pooling removes every orientation sensitivity and thus exhibits relatively lower performance.

\noindent\textbf{Parameter Efficiency.} Our method leverages rotation-equivariant convolution across all layers, from the backbone to the head, to ensure full rotation-equivariance. This maximizes the benefits of weight sharing, allowing us to outperform existing models with only 16\% of the parameters. As seen in \cref{tab:param}, even when compared to ReDet, which utilizes the same rotation-equivariant backbone, FRED has only 18\% of the size.

\begin{table}[t]
   \input{tables/parameter.tex}
    \caption{\textbf{Comparison of model size on DOTA-v1.5}}
   \label{tab:param}
\end{table}

\noindent\textbf{Observations on point set directionality.} Originally, rotation-equivariance is defined for various rotations of a single instance and is not applied to different instances. If rotation-equivariance also operates between multiple objects of the same class, we expect FRED to understand their relative orientations without any supervision. We observed that FRED captures the relative pose between objects during training (see \cref{fig:figure5}), even though its performance is lower than fully trained model. This tendency tends to diminish during the training process and only weakly appears in the converged model. We hypothesize that rotation-equivariance might coarsely cluster objects with similar shapes, sizes, and colors in the early stages of training, naturally making the model aware of their poses. However, since objects within the same class can have varied distributions, those pose sensitivity derived from coarse rotation-equivariance fades away. Nevertheless, this distinctly indicates that FRED engages in non-axis aligned feature learning. In contrast, the previous point set based methods produce predictions solely based on the bounding box distribution, agnostic to the object's direction. Even if they are not aligned to the horizontal or vertical axis, the predictions appear to be aligned to a fixed axis inherent to each point. We may leave this intriguing behavior of FRED as our future work.

\section{Conclusion}
In this paper, we introduced a fully rotation-equivariant object detector for aerial images. Our novel rotation-equivariant deformable convolution blocks offer improved alignment for spatial region and orientation with rotation-equivariant receptive fields. Moreover, through rotation-invariant orientation alignment, we strengthened the correlation between classification and localization. Not only does our model display rotation robustness, parameter efficiency, and promising quantitative results, but it also hints at the potential for unsupervised pose estimation.

\section{Acknowledgments}
This research was supported by a grant of the Korea Health Technology R\&D Project through the Korea Health Industry Development Institute (KHIDI), funded by the Ministry of Health and Welfare, Republic of Korea (grant number : HI20C1234).

\appendix

\bibliography{aaai24}

\end{document}

%% file: tables/dota_sota.tex
\definecolor{Gray}{gray}{0.87} 
\newcolumntype{g}{>{\columncolor{Gray}}c}

\begin{center}
\resizebox{\textwidth}{!}{%
\begin{tabular}{llccccccccccccccc|g}
\toprule
Methods                                     & Backbone      & PL        & BD        & BR        & GTF       & SV        & LV        & SH        & TC        & BC        & ST        & SBF       & RA        & HA        & SP        & HC        & mAP   \\
\midrule
\textbf{Anchor-based Methods:} \\
RoI-Transformer\citep{roitransformer}       & Res101-FPN    & 88.64     & 78.52     & 43.44     & 75.92     & 68.81     & 73.68     & 83.59     & 90.74     & 77.27     & 81.46     & 58.39     & 53.54     & 62.83     & 58.93     & 47.67     & 69.56 \\ 
FAOD\citep{li2019FAOD}                      & Res101-FPN    & {\textbf{90.21}}     & 79.58     & 45.49     & {\underline{76.41}}     & 73.18     & 68.27     & 79.56     & 90.83     & 83.40     & 84.68     & 53.40     & 65.42     & {\underline{74.17}}     & 69.69     & \underline{64.86}     & 73.28 \\
SCRDet\citep{scrdet}                        & Res101-FPN     & {\underline{89.98}}	    & 80.65	    & 52.09	    & 68.36	    & 68.36	    & 60.32	    & 72.41	    & 90.85     & \textbf{87.94} 	& \textbf{86.86}	    & \textbf{65.02}	    & \underline{66.68}	    & 66.25	    & 68.24	    & \textbf{65.21}	    & 72.61 \\
Gliding Vertex\citep{xu2020gliding}         & Res101-FPN    & 89.64     & {\textbf{85.00}}     & \underline{52.26}     & {\textbf{77.34}}     & 73.01     & 73.14     & 86.82     & 90.74     & 79.02     & {\underline{86.81}}     & 59.55     & {\textbf{70.91}}     & 72.94     & \underline{70.86}     & 57.32     & \underline{75.02} \\ 
R$^3$Det\citep{yang2021r3det}               & Res101-FPN    & 88.76     & \underline{83.09}     & 50.91     & 67.27     & 76.23     & \underline{80.39}     & 86.72     & 90.78     & 84.68     & 83.24     & 61.98     & 61.35     & 66.91     & 70.63     & 53.94     & 73.79 \\
S$^2$A-Net\citep{han2021s2anet}             & Res50-FPN     & 89.11     & 82.84     & 48.37     & 71.11     & \underline{78.11}     & 78.39     & \underline{87.25}     & 90.83     & 84.90     & 85.64     & 60.36     & 62.60     & 65.26     & 69.13     & 57.94     & 74.12 \\
DAL\citep{ming2021dynamic}                  & Res101-FPN    & 88.61     & 79.69     & 46.27     & 70.37     & 65.89     & 76.10     & 78.53     & 90.84     & 79.98     & 78.41     & 58.71     & 62.02     & 69.23     & {\underline{71.32}}     & 60.65     & 71.78 \\
KFIoU(R$^3$Det)\citep{yang2022kfiou}        & Res50-FPN     & 89.05     & 75.17     & 49.04     & 69.67     & 78.06     & 75.46     & 86.69     & {\textbf{90.90}}     & 83.65     & 84.48     & \underline{62.21}     & 62.87     & 66.72     & 65.95     & 50.20     & 72.68 \\
ReDet\citep{han2021ReDet}                   & ReRes50-ReFPN & 88.79     & 82.64     & {\textbf{53.97}}     & 74.00     & {\textbf{78.13}}     & {\textbf{84.06}}     & {\textbf{88.04}}     & {\underline{90.89}}     & {\underline{87.78}}     & 85.75     & 61.76     & 60.39     & {\textbf{75.96}}     & 68.07     & 63.59     & \textbf{76.25} \\ 
\midrule
\textbf{Anchor-free Methods:} \\
PIoU~\citep{chen2020piou}                   & DLA34         & 80.90     & 69.70     & 24.10     & 60.20     & 38.30     & 64.40     & 64.80     & {\textbf{90.90}}     & 77.20     & 70.40     & 46.50     & 37.10     & 57.10     & 61.90     & 64.00     & 60.50 \\
O$^2$-DNet\citep{wei2020o2det}              & H104          & 89.31     & 82.14     & 47.33     & 61.21     & 71.32     & 74.03     & 78.62     & 90.76     & 82.23     & 81.36     & 60.93     & 60.17     & 58.21     & 66.98     & 61.03     & 71.04 \\
DRN$^*$\citep{pan2020drn}                   & H104          & {\textbf{89.71}}     & 82.34     & 47.22     & 64.10     & 76.22     & 74.43     & 85.84     & 90.57     & {\underline{86.18}}     & 84.89     & 57.65     & 61.93     & 69.30     & 69.63     & 58.48     & 73.23 \\ 
CFA\citep{guo2021CFA}                       & Res101-FPN    & 89.26     & 81.72     & 51.81     & 67.17     & {\underline{79.99}}     & {\underline{78.25}}     & 84.46     & 90.77     & 83.40     & {\underline{85.54}}     & 54.86     & {\underline{67.75}}     & {\underline{73.04}}     & 70.24     & {\underline{64.96}}     & 75.05 \\
RSDet++\citep{qian2022rsdet++}              & Res152-FPN    & 86.80     & {\underline{82.70}}     & {\textbf{54.60}}     & {\underline{71.70}}     & 76.00     & 71.20     & 83.50     & 87.40     & 83.40     & 85.30     & {\textbf{72.40}}     & 62.90     & 70.90     & {\underline{72.00}}     & {\textbf{70.40}}     & 75.40 \\
Oriented RepPoints\citep{li2022oriented}    & Res50-FPN     & 87.02     & {\textbf{83.17}}     & {\underline{54.13}}     & 71.16     & {\textbf{80.18}}     & {\textbf{78.40}}     & {\underline{87.28}}     & {\textbf{90.90}}     & 85.97     & {\textbf{86.25}}     & 59.90     & {\textbf{70.49}}     & {\textbf{73.53}}     & {\textbf{72.27}}     & 58.97     & {\textbf{75.97}} \\
\rowcolor{Gray}
FRED (Ours)                                        & ReRes50-ReFPN & {\underline{89.37}}     & 82.12     & 50.84     & {\textbf{73.89}}     & 77.58     & 77.38     & {\textbf{87.51}}     & {\underline{90.82}}     & {\textbf{86.30}}     & 84.25     & {\underline{62.54}}     & 65.10     & 72.65     & 69.55     & 63.41     & {\underline{75.56}} \\
\bottomrule
\end{tabular}%
}
\end{center}

%% file: tables/dota_v15_sota.tex
\definecolor{Gray}{gray}{0.87} 
\newcolumntype{g}{>{\columncolor{Gray}}c}

\begin{center}
\resizebox{\textwidth}{!}{%
\begin{tabular}{lcccccccccccccccc|g}
\toprule
Methods                                     & PL    & BD    & BR    & GTF   & SV    & LV    & SH    & TC    & BC    & ST    & SBF   & RA    & HA    & SP    & HC    & CC    & mAP   \\ 
\midrule
RetinaNet-O\citep{lin2017focal}             & 71.43 & 77.64 & 42.12 & 64.65 & 44.53 & 56.79 & 73.31 & {\underline{90.84}} & 76.02 & 59.96 & 46.95 & 69.24 & 59.65 & 64.52 & 48.06 & 0.83  & 59.16 \\
Mask R-CNN\citep{he2017mask}                & 76.84 & 73.51 & 49.90 & 57.80 & 51.31 & 71.34 & 79.75 & 90.46 & 74.21 & 66.07 & 46.21 & 70.61 & 63.07 & 64.46 & {\underline{57.81}} & 9.42  & 62.67 \\
HTC\citep{chen2019hybrid}                   & 77.80 & 73.67 & 51.40 & 63.99 & 51.54 & 73.31 & 80.31 & 90.48 & 75.12 & 67.34 & 48.51 & 70.63 & 64.84 & 64.48 & 55.87 & 5.15  & 63.40 \\
RoI-Transformer\citep{roitransformer}       & 71.70 & {\underline{82.70}} & {\textbf{53.00}} & {\underline{71.50}} & 51.30 & 74.60 & 80.60 & 90.40 & 78.00 & 68.30 & 53.10 & {\textbf{73.40}} & {\textbf{73.90}} & 65.60 & 56.90 & 3.00  & 65.50 \\
ReDet\citep{han2021ReDet}                   & {\underline{79.20}} & {\textbf{82.81}} & 51.92 & 71.41 & 52.38 & {\textbf{75.73}} & 80.92 & 90.83 & 75.81 & 68.64 & 49.29 & 72.03 & {\underline{73.36}} & {\textbf{70.55}} & {\textbf{63.33}} & {\underline{11.53}} & 66.86 \\
Oriented RepPoints\citep{li2022oriented}    & 75.52 & 82.60 & 51.24 & 70.21 & {\underline{57.81}} & 73.82 & {\textbf{86.25}} & {\textbf{90.86}} & {\underline{78.30}} & {\textbf{76.47}} & {\textbf{53.61}} & 72.78 & 66.68 & 69.48 & 53.66 & 11.09 & {\underline{66.90}} \\
\rowcolor{Gray}
FRED (Ours)                                  & {\textbf{79.60}} & 81.44 & {\underline{52.60}} & {\textbf{72.57}} & {\textbf{58.07}} & {\underline{74.82}} & {\underline{86.12}} & 90.81 & {\textbf{82.13}} & {\underline{74.84}} & {\underline{53.37}} & {\underline{72.93}} & 69.51 & {\underline{69.91}} & 54.82 & {\textbf{19.27}} & {\textbf{68.30}} \\
\bottomrule
\end{tabular}%
}
\end{center}

%% file: tables/ablation.tex
\definecolor{Gray}{gray}{0.87} 
\newcolumntype{g}{>{\columncolor{Gray}}c}

\begin{center}
\resizebox{\linewidth}{!}{
    \begin{tabular}{lccccc}
        \toprule
         & \multicolumn{5}{c}{Method} \\
        \midrule
        \textbf{Vector-field} & - & - & \checkmark & \checkmark & \checkmark \\
        \textbf{GroupPool} & - & \checkmark & \checkmark & - & - \\
        \textbf{OA} & - & - & - & \checkmark & \checkmark \\
        \textbf{Edge constraint} & - & - & - & - & \checkmark \\
        \midrule
        \textbf{mAP} & 73.32 & 73.91 & 74.69 & 75.35 & 75.56 \\
        \bottomrule
        \end{tabular}
}
\end{center}

%% file: tables/parameter.tex
\definecolor{Gray}{gray}{0.87} 
\newcolumntype{g}{>{\columncolor{Gray}}c}

\begin{center}
\resizebox{\linewidth}{!}{
\begin{tabular}{lccc}
\toprule
Methods               & Backbone      & mAP   & Size (MB) \\
\midrule
Oriented RepPoints    & Res50-FPN     & 66.90 & 140.78 \\
ReDet                 & ReRes50-ReFPN & 66.86 & 124.97 \\
\rowcolor{Gray}
FRED (Ours)            & ReRes50-ReFPN & \textbf{68.30} & \textbf{22.28} \\
\bottomrule
\end{tabular}%
}
\end{center}

%% file: FRED_CameraReady.bbl
\begin{thebibliography}{34}
\providecommand{\natexlab}[1]{#1}

\bibitem[{Cesa, Lang, and Weiler(2021)}]{cesa2021program}
Cesa, G.; Lang, L.; and Weiler, M. 2021.
\newblock A program to build E (N)-equivariant steerable CNNs.
\newblock In \emph{International Conference on Learning Representations}.

\bibitem[{Chen et~al.(2019)Chen, Pang, Wang, Xiong, Li, Sun, Feng, Liu, Shi, Ouyang et~al.}]{chen2019hybrid}
Chen, K.; Pang, J.; Wang, J.; Xiong, Y.; Li, X.; Sun, S.; Feng, W.; Liu, Z.; Shi, J.; Ouyang, W.; et~al. 2019.
\newblock Hybrid task cascade for instance segmentation.
\newblock In \emph{Proceedings of the IEEE/CVF conference on computer vision and pattern recognition}, 4974--4983.

\bibitem[{Chen et~al.(2020)Chen, Chen, Lin, See, Yu, Ke, and Yang}]{chen2020piou}
Chen, Z.; Chen, K.; Lin, W.; See, J.; Yu, H.; Ke, Y.; and Yang, C. 2020.
\newblock Piou loss: Towards accurate oriented object detection in complex environments.
\newblock In \emph{Computer Vision--ECCV 2020: 16th European Conference, Glasgow, UK, August 23--28, 2020, Proceedings, Part V 16}, 195--211. Springer.

\bibitem[{Cohen and Welling(2016)}]{cohen2016group}
Cohen, T.; and Welling, M. 2016.
\newblock Group equivariant convolutional networks.
\newblock In \emph{International conference on machine learning}, 2990--2999. PMLR.

\bibitem[{Ding et~al.(2019)Ding, Xue, Long, Xia, and Lu}]{roitransformer}
Ding, J.; Xue, N.; Long, Y.; Xia, G.-S.; and Lu, Q. 2019.
\newblock Learning RoI transformer for oriented object detection in aerial images.
\newblock In \emph{Proceedings of the IEEE/CVF Conference on Computer Vision and Pattern Recognition}, 2849--2858.

\bibitem[{Ding et~al.(2021)Ding, Xue, Xia, Bai, Yang, Yang, Belongie, Luo, Datcu, Pelillo, and Zhang}]{9560031}
Ding, J.; Xue, N.; Xia, G.-S.; Bai, X.; Yang, W.; Yang, M.; Belongie, S.; Luo, J.; Datcu, M.; Pelillo, M.; and Zhang, L. 2021.
\newblock Object Detection in Aerial Images: A Large-Scale Benchmark and Challenges.
\newblock \emph{IEEE Transactions on Pattern Analysis and Machine Intelligence}, 1--1.

\bibitem[{Everingham et~al.(2010)Everingham, Van~Gool, Williams, Winn, and Zisserman}]{everingham2010pascal}
Everingham, M.; Van~Gool, L.; Williams, C.~K.; Winn, J.; and Zisserman, A. 2010.
\newblock The pascal visual object classes (voc) challenge.
\newblock \emph{International journal of computer vision}, 88: 303--338.

\bibitem[{Guo et~al.(2021{\natexlab{a}})Guo, Liu, Zhang, Jiao, Ji, and Ye}]{guo2021beyond}
Guo, Z.; Liu, C.; Zhang, X.; Jiao, J.; Ji, X.; and Ye, Q. 2021{\natexlab{a}}.
\newblock Beyond bounding-box: Convex-hull feature adaptation for oriented and densely packed object detection.
\newblock In \emph{Proceedings of the IEEE/CVF conference on Computer Vision and Pattern Recognition}, 8792--8801.

\bibitem[{Guo et~al.(2021{\natexlab{b}})Guo, Liu, Zhang, Jiao, Ji, and Ye}]{guo2021CFA}
Guo, Z.; Liu, C.; Zhang, X.; Jiao, J.; Ji, X.; and Ye, Q. 2021{\natexlab{b}}.
\newblock Beyond bounding-box: Convex-hull feature adaptation for oriented and densely packed object detection.
\newblock In \emph{Proceedings of the IEEE/CVF conference on Computer Vision and Pattern Recognition}, 8792--8801.

\bibitem[{Gupta, Arya, and Gavves(2021)}]{gupta2021rotation}
Gupta, D.~K.; Arya, D.; and Gavves, E. 2021.
\newblock Rotation equivariant siamese networks for tracking.
\newblock In \emph{Proceedings of the IEEE/CVF Conference on Computer Vision and Pattern Recognition}, 12362--12371.

\bibitem[{Han et~al.(2021{\natexlab{a}})Han, Ding, Li, and Xia}]{han2021s2anet}
Han, J.; Ding, J.; Li, J.; and Xia, G.-S. 2021{\natexlab{a}}.
\newblock Align deep features for oriented object detection.
\newblock \emph{IEEE Transactions on Geoscience and Remote Sensing}, 60: 1--11.

\bibitem[{Han et~al.(2021{\natexlab{b}})Han, Ding, Xue, and Xia}]{han2021ReDet}
Han, J.; Ding, J.; Xue, N.; and Xia, G.-S. 2021{\natexlab{b}}.
\newblock Redet: A rotation-equivariant detector for aerial object detection.
\newblock In \emph{Proceedings of the IEEE/CVF Conference on Computer Vision and Pattern Recognition}, 2786--2795.

\bibitem[{He et~al.(2017)He, Gkioxari, Doll{\'a}r, and Girshick}]{he2017mask}
He, K.; Gkioxari, G.; Doll{\'a}r, P.; and Girshick, R. 2017.
\newblock Mask r-cnn.
\newblock In \emph{Proceedings of the IEEE international conference on computer vision}, 2961--2969.

\bibitem[{Lee et~al.(2023)Lee, Kim, Kim, and Cho}]{lee2023learning}
Lee, J.; Kim, B.; Kim, S.; and Cho, M. 2023.
\newblock Learning Rotation-Equivariant Features for Visual Correspondence.
\newblock In \emph{Proceedings of the IEEE/CVF Conference on Computer Vision and Pattern Recognition}, 21887--21897.

\bibitem[{Li et~al.(2019)Li, Xu, Cui, Wang, Zhang, and Yang}]{li2019FAOD}
Li, C.; Xu, C.; Cui, Z.; Wang, D.; Zhang, T.; and Yang, J. 2019.
\newblock Feature-attentioned object detection in remote sensing imagery.
\newblock In \emph{2019 IEEE international conference on image processing (ICIP)}, 3886--3890. IEEE.

\bibitem[{Li et~al.(2022)Li, Chen, Hu, and Zhu}]{li2022oriented}
Li, W.; Chen, Y.; Hu, K.; and Zhu, J. 2022.
\newblock Oriented reppoints for aerial object detection.
\newblock In \emph{Proceedings of the IEEE/CVF conference on computer vision and pattern recognition}, 1829--1838.

\bibitem[{Lin et~al.(2017)Lin, Goyal, Girshick, He, and Doll{\'a}r}]{lin2017focal}
Lin, T.-Y.; Goyal, P.; Girshick, R.; He, K.; and Doll{\'a}r, P. 2017.
\newblock Focal loss for dense object detection.
\newblock In \emph{Proceedings of the IEEE international conference on computer vision}, 2980--2988.

\bibitem[{Ming et~al.(2021)Ming, Zhou, Miao, Zhang, and Li}]{ming2021dynamic}
Ming, Q.; Zhou, Z.; Miao, L.; Zhang, H.; and Li, L. 2021.
\newblock Dynamic anchor learning for arbitrary-oriented object detection.
\newblock In \emph{Proceedings of the AAAI Conference on Artificial Intelligence}, volume~35, 2355--2363.

\bibitem[{Pan et~al.(2020)Pan, Ren, Sheng, Dong, Yuan, Guo, Ma, and Xu}]{pan2020drn}
Pan, X.; Ren, Y.; Sheng, K.; Dong, W.; Yuan, H.; Guo, X.; Ma, C.; and Xu, C. 2020.
\newblock Dynamic refinement network for oriented and densely packed object detection.
\newblock In \emph{Proceedings of the IEEE/CVF Conference on Computer Vision and Pattern Recognition}, 11207--11216.

\bibitem[{Qian et~al.(2021)Qian, Yang, Peng, Yan, and Guo}]{qian2021rsdet}
Qian, W.; Yang, X.; Peng, S.; Yan, J.; and Guo, Y. 2021.
\newblock Learning modulated loss for rotated object detection.
\newblock In \emph{Proceedings of the AAAI conference on artificial intelligence}, volume~35, 2458--2466.

\bibitem[{Qian et~al.(2022)Qian, Yang, Peng, Zhang, and Yan}]{qian2022rsdet++}
Qian, W.; Yang, X.; Peng, S.; Zhang, X.; and Yan, J. 2022.
\newblock RSDet++: Point-based modulated loss for more accurate rotated object detection.
\newblock \emph{IEEE Transactions on Circuits and Systems for Video Technology}, 32(11): 7869--7879.

\bibitem[{Rezatofighi et~al.(2019)Rezatofighi, Tsoi, Gwak, Sadeghian, Reid, and Savarese}]{rezatofighi2019convexiou}
Rezatofighi, H.; Tsoi, N.; Gwak, J.; Sadeghian, A.; Reid, I.; and Savarese, S. 2019.
\newblock Generalized intersection over union: A metric and a loss for bounding box regression.
\newblock In \emph{Proceedings of the IEEE/CVF conference on computer vision and pattern recognition}, 658--666.

\bibitem[{Romero et~al.(2020)Romero, Bekkers, Tomczak, and Hoogendoorn}]{romero2020attentive}
Romero, D.; Bekkers, E.; Tomczak, J.; and Hoogendoorn, M. 2020.
\newblock Attentive group equivariant convolutional networks.
\newblock In \emph{International Conference on Machine Learning}, 8188--8199. PMLR.

\bibitem[{Veeling et~al.(2018)Veeling, Linmans, Winkens, Cohen, and Welling}]{veeling2018rotation}
Veeling, B.~S.; Linmans, J.; Winkens, J.; Cohen, T.; and Welling, M. 2018.
\newblock Rotation equivariant CNNs for digital pathology.
\newblock In \emph{Medical Image Computing and Computer Assisted Intervention--MICCAI 2018: 21st International Conference, Granada, Spain, September 16-20, 2018, Proceedings, Part II 11}, 210--218. Springer.

\bibitem[{Wei et~al.(2020)Wei, Zhang, Chang, Li, Wang, and Sun}]{wei2020o2det}
Wei, H.; Zhang, Y.; Chang, Z.; Li, H.; Wang, H.; and Sun, X. 2020.
\newblock Oriented objects as pairs of middle lines.
\newblock \emph{ISPRS Journal of Photogrammetry and Remote Sensing}, 169: 268--279.

\bibitem[{Weiler and Cesa(2019)}]{weiler2019general}
Weiler, M.; and Cesa, G. 2019.
\newblock General e (2)-equivariant steerable cnns.
\newblock \emph{Advances in neural information processing systems}, 32.

\bibitem[{Xia et~al.(2018)Xia, Bai, Ding, Zhu, Belongie, Luo, Datcu, Pelillo, and Zhang}]{DOTA}
Xia, G.-S.; Bai, X.; Ding, J.; Zhu, Z.; Belongie, S.; Luo, J.; Datcu, M.; Pelillo, M.; and Zhang, L. 2018.
\newblock DOTA: A Large-Scale Dataset for Object Detection in Aerial Images.
\newblock In \emph{The IEEE Conference on Computer Vision and Pattern Recognition (CVPR)}.

\bibitem[{Xu et~al.(2020)Xu, Fu, Wang, Wang, Chen, Xia, and Bai}]{xu2020gliding}
Xu, Y.; Fu, M.; Wang, Q.; Wang, Y.; Chen, K.; Xia, G.-S.; and Bai, X. 2020.
\newblock Gliding vertex on the horizontal bounding box for multi-oriented object detection.
\newblock \emph{IEEE transactions on pattern analysis and machine intelligence}, 43(4): 1452--1459.

\bibitem[{Yang et~al.(2021)Yang, Yan, Feng, and He}]{yang2021r3det}
Yang, X.; Yan, J.; Feng, Z.; and He, T. 2021.
\newblock R3det: Refined single-stage detector with feature refinement for rotating object.
\newblock In \emph{Proceedings of the AAAI conference on artificial intelligence}, volume~35, 3163--3171.

\bibitem[{Yang et~al.(2019{\natexlab{a}})Yang, Yang, Yan, Zhang, Zhang, Guo, Sun, and Fu}]{scrdet}
Yang, X.; Yang, J.; Yan, J.; Zhang, Y.; Zhang, T.; Guo, Z.; Sun, X.; and Fu, K. 2019{\natexlab{a}}.
\newblock Scrdet: Towards more robust detection for small, cluttered and rotated objects.
\newblock In \emph{Proceedings of the IEEE/CVF international conference on computer vision}, 8232--8241.

\bibitem[{Yang et~al.(2022)Yang, Zhou, Zhang, Yang, Wang, Yan, Zhang, and Tian}]{yang2022kfiou}
Yang, X.; Zhou, Y.; Zhang, G.; Yang, J.; Wang, W.; Yan, J.; Zhang, X.; and Tian, Q. 2022.
\newblock The KFIoU loss for rotated object detection.
\newblock \emph{arXiv preprint arXiv:2201.12558}.

\bibitem[{Yang et~al.(2019{\natexlab{b}})Yang, Liu, Hu, Wang, and Lin}]{yang2019reppoints}
Yang, Z.; Liu, S.; Hu, H.; Wang, L.; and Lin, S. 2019{\natexlab{b}}.
\newblock Reppoints: Point set representation for object detection.
\newblock In \emph{Proceedings of the IEEE/CVF international conference on computer vision}, 9657--9666.

\bibitem[{Zhou et~al.(2022)Zhou, Yang, Zhang, Wang, Liu, Hou, Jiang, Liu, Yan, Lyu et~al.}]{zhou2022mmrotate}
Zhou, Y.; Yang, X.; Zhang, G.; Wang, J.; Liu, Y.; Hou, L.; Jiang, X.; Liu, X.; Yan, J.; Lyu, C.; et~al. 2022.
\newblock Mmrotate: A rotated object detection benchmark using pytorch.
\newblock In \emph{Proceedings of the 30th ACM International Conference on Multimedia}, 7331--7334.

\bibitem[{Zhu et~al.(2019)Zhu, Hu, Lin, and Dai}]{zhu2019deformable}
Zhu, X.; Hu, H.; Lin, S.; and Dai, J. 2019.
\newblock Deformable convnets v2: More deformable, better results.
\newblock In \emph{Proceedings of the IEEE/CVF conference on computer vision and pattern recognition}, 9308--9316.

\end{thebibliography}
